%% file: main.tex
\documentclass[10pt,twocolumn,letterpaper]{article}

\usepackage[pagenumbers]{cvpr} % To force page numbers, e.g. for an arXiv version

\input{preamble}

\definecolor{cvprblue}{rgb}{0.21,0.49,0.74}
\usepackage[pagebackref,breaklinks,colorlinks,allcolors=cvprblue]{hyperref}

\def\confName{CVPR}
\def\confYear{2027}

\title{Beyond Which: Where, How, and When in Vision-Language-Action Models}
\author{Yunhe Zhang, Long Gao, Yuedong Fu, Gang He, Li Xu, Chang Wu, Songyuheng Gao\\
Xidian University\\
{\tt\small yunhezhang@stu.xidian.edu.cn}
}

\begin{document}
\maketitle
\input{sec/0_abstract}
\input{sec/1_introduction}

\input{sec/2_related_work}
\input{sec/3_method}
\input{sec/4_experiments}
\input{sec/5_conclusion}
\input{sec/6_llm}
{
    \small
    \bibliographystyle{ieeenat_fullname}
    \bibliography{main}
}


\end{document}

%% file: preamble.tex
\usepackage{array}
\usepackage{colortbl}
\usepackage{booktabs}
\usepackage{tabularx}
\usepackage{bm}
\usepackage{multirow}

\definecolor{oursbg}{RGB}{240,240,240}

\newcommand{\method}{\mbox{\ensuremath{\mathcal{W}^{3}\mathcal{H}}\textnormal{-VLA}}\xspace}
\newcommand{\piZero}{\ensuremath{\pi_{0}}\xspace}
\newcommand{\piHalf}{\ensuremath{\pi_{0.5}}\xspace}
\newcommand{\cmark}{\ensuremath{\checkmark}}
\newcommand{\xmark}{\ensuremath{\times}}
\newcommand{\best}[1]{\textbf{#1}}
\newcommand{\second}[1]{\underline{#1}}

\DeclareMathOperator{\CrossAttn}{Cross-Attn}
\DeclareMathOperator{\Concat}{Concat}
\DeclareMathOperator{\FFN}{FFN}
\DeclareMathOperator{\MLP}{MLP}

%% file: sec/0_abstract.tex
\begin{abstract}
Vision-language-action (VLA) models typically rely on vision-language model (VLM) features as the primary perceptual source for action prediction, showing strong generalization across diverse robotic manipulation tasks. Although VLM features can identify which entities are relevant to action prediction, precise manipulation also requires understanding where task-relevant entities are located within the spatial layout, how their geometric structure constrains physical interaction, and when spatial or geometric information should make a greater contribution to action prediction. To this end, we introduce \method to organize VLA perception and action prediction around \textit{which}, \textit{where}, \textit{how}, and \textit{when}. Specifically, \method retains VLM-based semantic entity perception (\textit{which}) and further introduces spatial distance perception (\textit{where}) and geometric structure perception (\textit{how}) with adaptive spatial-geometric guidance (\textit{when}) to enrich the perceptual information available for action prediction. Experiments across real-world settings and simulation benchmarks demonstrate the effectiveness and superior performance of \method.

% The ABSTRACT is to be in fully justified italicized text, at the top of the left-hand column, below the author and affiliation information.
% Use the word ``Abstract'' as the title, in 12-point Times, boldface type, centered relative to the column, initially capitalized.
% The abstract is to be in 10-point, single-spaced type.
% Leave two blank lines after the Abstract, then begin the main text.
% Look at previous \confName abstracts to get a feel for style and length.
\end{abstract}

%% file: sec/1_introduction.tex
\section{Introduction}
\label{sec:introduction}

Robot policies optimized for specific tasks often suffer from limited generalization~\cite{kim2024openvla}. Meanwhile, recent advances in large-scale vision-language models (VLMs) have demonstrated strong capabilities in multi-modal learning and reasoning~\cite{achiam2023gpt, beyer2024paligemma}. The advances provide a strong basis for visual perception and language understanding, enabling robot policies to leverage broad vision-language knowledge for interpreting visual observations and language instructions across the physical world. Building upon these, vision-language-action (VLA) models combine a pre-trained VLM for perceptual and semantic understanding with an action prediction module that maps observations and instructions to robot actions, enabling robot policies to generalize across diverse tasks~\cite{zitkovich2023rt, kim2024openvla, black2024pi_0, intelligence2025pi_, zheng2026x, luo2026simvla}.

\begin{figure}[t]
    \centering
    \includegraphics[width=0.95\columnwidth]{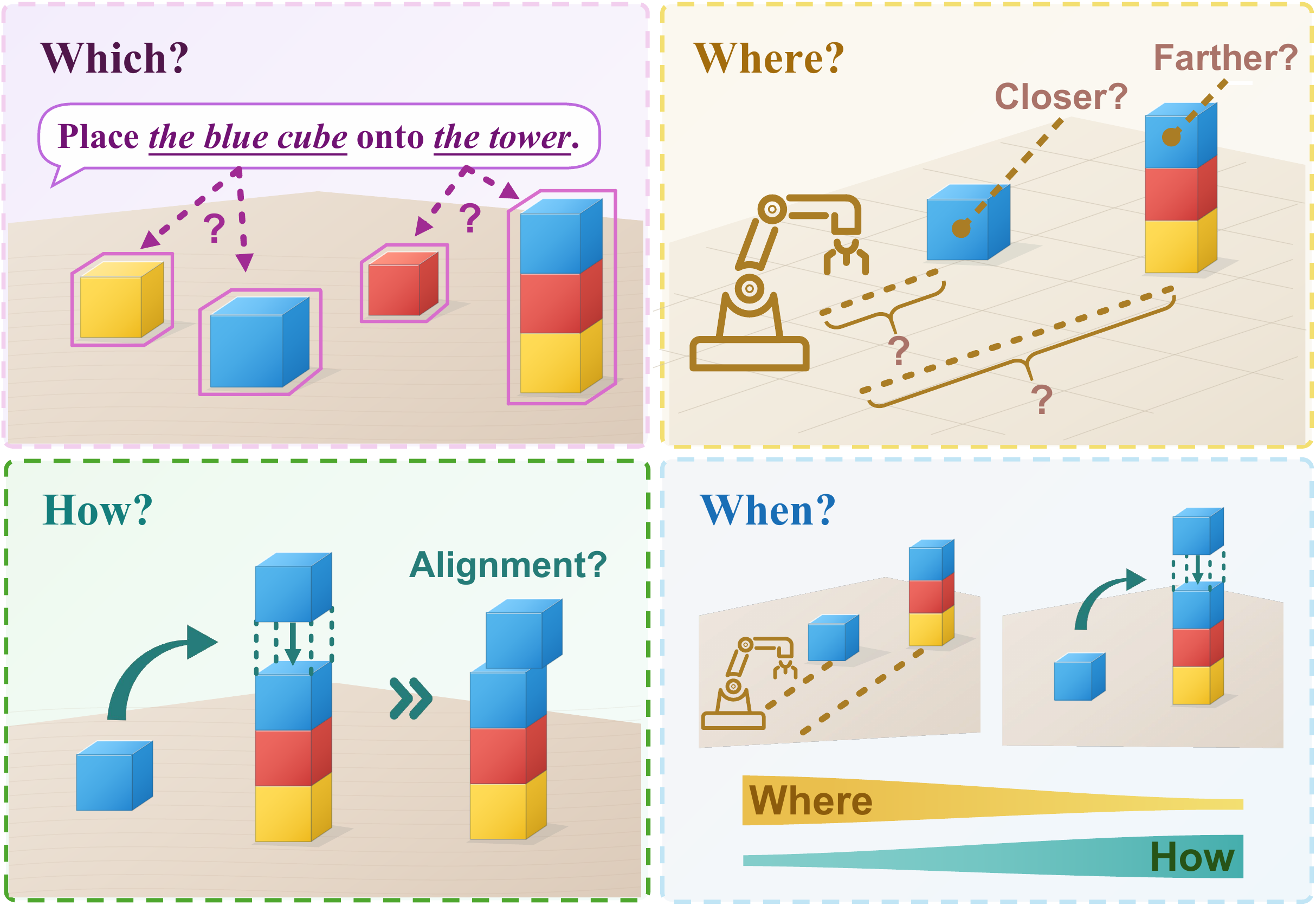}
    \caption{\method organizes VLA perception and action prediction around \textit{which}, \textit{where}, \textit{how}, and \textit{when}.}
    \label{fig:insight}
\end{figure}

Despite the strong generalization, most VLA models still rely on the VLM-based 2D visual perception, resulting in a primary focus on the semantic information about the entities relevant to the action prediction. Recent efforts have begun to enrich perception with spatial depth information~\cite{qu2025spatialvla} and 3D scene representations~\cite{zhen20243d}. Although these advances show the benefits of richer perception, they also motivate a broader question: \textbf{\textit{what perceptual information is required to support precise action prediction in a VLA model?}}

Consider the task of placing a blue cube surrounded by other cubes onto a cube tower. Given 2D visual observations and a language instruction, the VLM provides semantic information for identifying \textbf{\underline{\textit{which}}} cube to manipulate. Semantic information therefore remains essential throughout the manipulation process to ensure that action prediction remains grounded in relevant entities. Beyond this, precise action prediction also depends on knowing \textbf{\underline{\textit{where}}} the cube, the tower, and the end effector are located relative to one another. Furthermore, the model must also account for \textbf{\underline{\textit{how}}} the shapes and local geometric details of the entities constrain grasping, alignment, and placement. During manipulation, it is also required to decide \textbf{\underline{\textit{when}}} action prediction should emphasize spatial relationships for localization or prioritize geometric constraints for grasping, alignment, and placement.

Accordingly, we argue that precise action prediction requires not only VLM-based semantic entity perception but also spatial distance perception, geometric structure perception, and adaptive spatial-geometric guidance, which together advance VLA perception beyond \textit{\textbf{Which}} with the consideration of \textbf{\textit{Where}}, \textbf{\textit{How}}, and \textbf{\textit{When}}.
 
As shown in~\cref{fig:insight}, we introduce \method, a novel framework that organizes VLA perception and action prediction around \textit{which}, \textit{where}, \textit{how}, and \textit{when}. For \textit{which}, \method follows the standard VLA paradigm, using a VLM to derive semantic entity perception from 2D visual observations and language instructions, thereby preserving the full semantic information as the primary task-relevant context throughout action prediction. In terms of \textit{where}, \method introduces spatial distance perception by incorporating task-relevant distance relationships and spatial layout features derived from a pre-trained depth estimation model \cite{yang2024depth}. About \textit{how}, \method enables geometric structure perception by capturing task-relevant geometric structure features from a pre-trained visual geometry model \cite{wang2026vggt}. For \textit{when}, \method integrates adaptive spatial-geometric guidance into action prediction by coordinating the contributions of spatial distance perception and geometric structure perception, conditioned on the current task semantic context and evolving action state. By unifying \textit{which}, \textit{where}, \textit{how}, and \textit{when}, \method complements semantic entity perception with spatial distance perception, geometric structure perception, and adaptive spatial-geometric guidance to support precise action prediction in the complex physical world.

To validate the effectiveness of \method, we conduct extensive experiments on 4 benchmarks, LIBERO~\cite{liu2023libero}, LIBERO-Plus~\cite{fei2025libero}, CALVIN~\cite{mees2022calvin}, LIBERO-PRO~\cite{zhou2025libero}, and 4 real-world manipulation tasks. The contributions of this work are summarized as follows.
\begin{itemize}
    \item We formulate VLA perception and action prediction in terms of \textit{which}, \textit{where}, \textit{how}, and \textit{when}, extending VLM-based semantic entity perception with spatial distance perception, geometric structure perception, and adaptive spatial-geometric guidance.
    \item Building on the formulation and the corresponding perceptual design, we propose \method, a novel VLA framework that incorporates a broader range of perceptual information into action prediction.
    \item We evaluate \method on 4 benchmarks and 4 real-world manipulation tasks, showing its effectiveness and superior performance in simulated and physical settings.
\end{itemize}

\begin{figure*}[t]
    \centering
    \includegraphics[width=\textwidth]{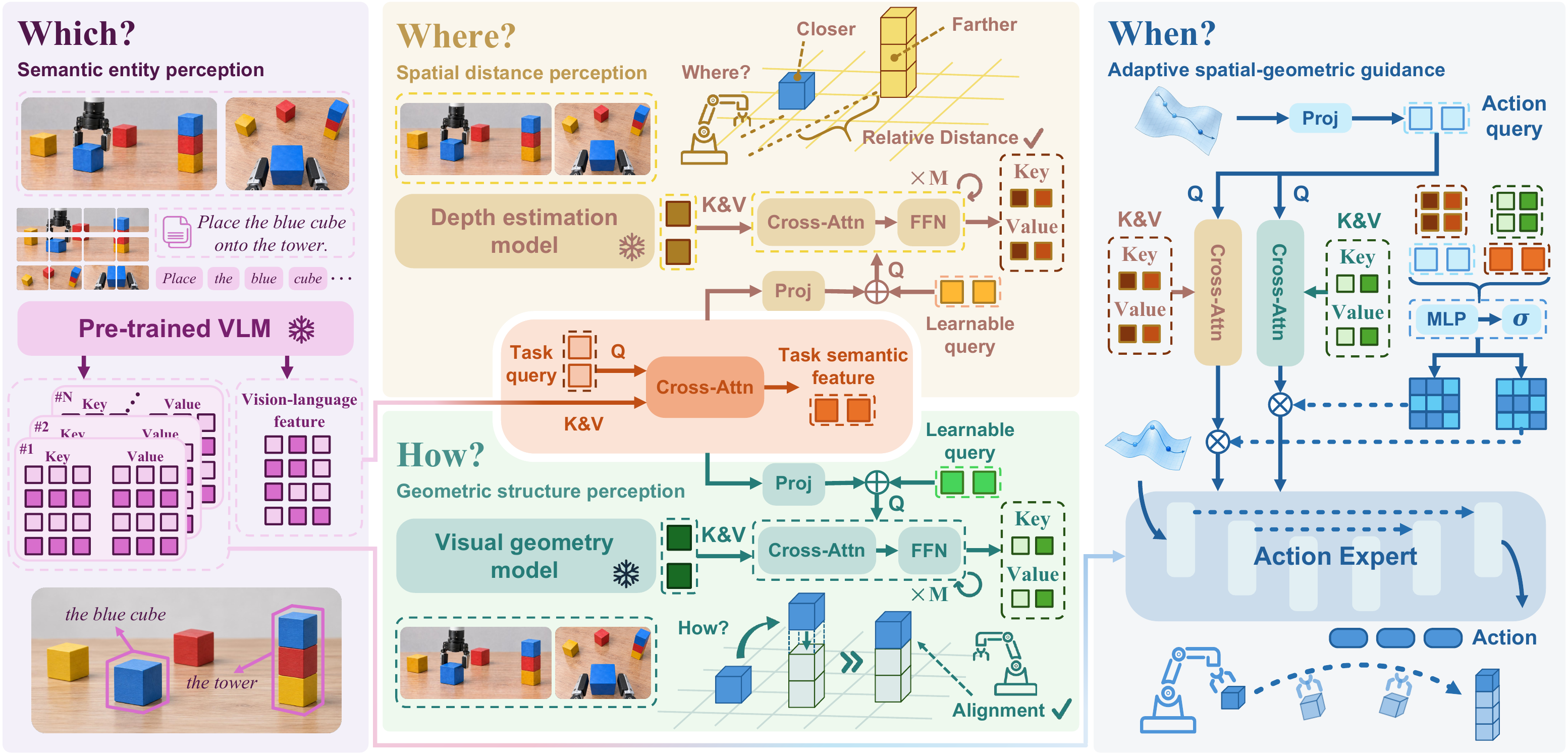}
    \caption{Overview of \method. The pre-trained VLM supplies semantic entity perception (\textit{which}). The spatial distance perception (\textit{where}) and geometric structure perception (\textit{how}) extract richer task-relevant perceptual information, and adaptive spatial-geometric guidance (\textit{when}) coordinates the contributions of the spatial distance and geometric structure perception.}
    \label{fig:pipeline}
\end{figure*}

%% file: sec/2_related_work.tex
\section{Related Work}
\label{sec:related}

\subsection{Vision-Language-Action Models}
VLA models build upon pre-trained VLMs to map visual observations and language instructions to robot actions~\cite{black2024pi_0, qu2025spatialvla, zheng2026x, luo2026simvla}. One line of research advances VLA models by scaling robot interaction data and vision-language pre-training. RT-1~\cite{brohan2022rt} demonstrates the benefits of large and diverse real-world robot data, and RT-2~\cite{zitkovich2023rt} transfers web-scale vision-language knowledge to action prediction. Open X-Embodiment~\cite{o2024open} aggregates cross-embodiment robot data, enabling the training of generalist policies~\cite{team2024octo, kim2024openvla}. More recently, \piHalf~\cite{intelligence2025pi_} integrates heterogeneous robot and web data to improve open-world generalization. Another line of research improves VLA action prediction through specialized policy architectures. RDT-1B~\cite{liu2025rdt} employs the diffusion model for continuous action prediction. Within the flow matching paradigm, \piZero~\cite{black2024pi_0} introduces an action expert for continuous action chunks, X-VLA~\cite{zheng2026x} incorporates soft prompts for cross-embodiment learning, and SimVLA~\cite{luo2026simvla} adopts a lightweight action transformer.

\subsection{Perception for Robotic Manipulation}
Research on robotic perception has progressively expanded from 2D semantic perception to spatial and geometric representations, and more recently to adaptive perceptual guidance.
Early methods primarily extract semantic features from 2D visual observations. CLIPort~\cite{shridhar2022cliport} combines VLM-derived semantic features with spatially precise action prediction, and RoboGround~\cite{huang2025roboground} predicts masks for relevant entities and placement regions to guide manipulation. Beyond 2D semantics, subsequent methods incorporate depth and 3D representations to capture physical structure features. PerAct~\cite{shridhar2023perceiver} encodes depth observations into a voxel grid, Act3D~\cite{gervet2023act3d} uses depth observations to lift pre-trained 2D features into a 3D field, and RVT~\cite{goyal2023rvt} aggregates features across virtual views for 3D-aware action prediction. Recent studies further explore adaptive perceptual guidance conditioned on task demands and action execution context. HyperTASR~\cite{sun2025hypertasr} adaptively modulates scene features according to task and execution progress, and ActiveVLA~\cite{liu2026activevla} dynamically selects 3D regions, viewpoints, and spatial resolutions for precise manipulation.

In contrast, \method unifies VLM-based semantic entity perception to identify \textit{which} entities are relevant to action prediction, spatial distance perception to establish \textit{where} relevant entities are located, geometric structure perception to capture \textit{how} the geometric details constrain physical interaction, and adaptive spatial-geometric guidance to determine \textit{when} the spatial distance or geometric structure perception should make a greater contribution to precise action prediction.

%% file: sec/3_method.tex
\section{Method}
\label{sec:method}

This section details how \method organizes VLA perception and action prediction. \cref{sec:preliminary} introduces the problem formulation and base VLA model, followed by semantic entity perception (\textit{which}) in \cref{sec:which}. Spatial distance perception (\textit{where}) and geometric structure perception (\textit{how}) are presented in \cref{sec:where_how}, and adaptive spatial-geometric guidance (\textit{when}) in \cref{sec:when}. Finally, \cref{sec:training_inference} describes the training and inference procedures.

\subsection{Preliminary}
\label{sec:preliminary}

\textbf{Problem formulation.}
At time step $t$, given the 2D visual observation $\mathbf{I}_t$ and a language instruction $\mathbf{L}$, the robot policy $\pi_\theta$ predicts a continuous action chunk $\mathbf{A}_t=[\mathbf{a}_t,\mathbf{a}_{t+1},\ldots,\mathbf{a}_{t+H-1}]$ over an action horizon of $H$ steps.
\begin{equation}
    \mathbf{A}_t = \pi_\theta\!\left(\mathbf{I}_t,\mathbf{L}\right),
    \label{eq:policy}
\end{equation}
% where $\mathbf{a}_{t+h}$ denotes the robot action at horizon step $h$. 

Although VLA models have shown generalization relying primarily on semantic entity perception derived from 2D visual observations, precise action prediction often requires additional forms of perception. Consequently, \method organizes VLA perception and action prediction beyond semantic entity perception (\textit{which}) by introducing spatial distance perception (\textit{where}), geometric structure perception (\textit{how}), and adaptive spatial-geometric guidance (\textit{when}).

\noindent\textbf{Base VLA model.}
\method adopts \piHalf~\cite{intelligence2025pi_} as the base model. \piHalf comprises a pre-trained PaliGemma VLM~\cite{beyer2024paligemma} and an action expert. The VLM encodes the 2D visual observations and language instruction into vision-language features, which the action expert leverages to predict continuous action chunks via conditional flow matching~\cite{lipman2022flow}.

\subsection{Which?}
\label{sec:which}

The VLM jointly encodes $\mathbf{I}_t$ and $\mathbf{L}$, producing the final vision-language feature $\mathbf{F}^{\mathrm{VLM}}$ and an $N$-layer vision-language key-value cache $\mathbf{C}^{\mathrm{VLM}}=\{(\mathbf{K}^{\mathrm{VLM}}_l,\mathbf{V}^{\mathrm{VLM}}_l)\}_{l=1}^{N}$.
\begin{equation}
    \left(\mathbf{F}^{\mathrm{VLM}},\mathbf{C}^{\mathrm{VLM}}\right)
    = f_{\mathrm{VLM}}\!\left(\mathbf{I}_t,\mathbf{L}\right).
    \label{eq:vlm}
\end{equation}

% The vision-language key-value cache provides semantic entity perception, and \method retains the resulting semantic information in full as the primary task-relevant context throughout action prediction. The final visual-language feature further serves as the task-relevant basis for extending VLA perception and action prediction from \textit{which} toward \textit{where}, \textit{how}, and \textit{when}.

The vision-language key-value cache provides semantic entity perception and is retained in full as the primary task-relevant context throughout action prediction. The final vision-language feature serves as the task-relevant basis for extending VLA perception beyond \textit{which} with the consideration of \textit{where}, \textit{how}, and \textit{when}.

\subsection{Where? and How?}
\label{sec:where_how}

Beyond semantic entity perception (\textit{which}), \method organizes spatial distance perception (\textit{where}) and geometric structure perception (\textit{how}) into two independent branches. The spatial distance perception branch employs a pre-trained depth estimation model to extract the spatial distance feature $\mathbf{F}^{\mathrm{Spa}}$ from $\mathbf{I}_t$, encoding distance relationships and spatial layout. In parallel, the geometric structure perception branch employs a pre-trained visual geometry model to derive the geometric structure feature $\mathbf{F}^{\mathrm{Geo}}$, capturing geometric details relevant to physical interaction constraints.

To enable task-relevant perception feature extraction in both perception branches, a task query $\mathbf{Q}^{\mathrm{Task}}$ first aggregates a task semantic feature $\mathbf{F}^{\mathrm{Task}}$ from $\mathbf{F}^{\mathrm{VLM}}$ as follows.
\begin{equation}
    \mathbf{F}^{\mathrm{Task}}
    = \CrossAttn^{\mathrm{Task}}\!\left(
        \mathbf{Q}^{\mathrm{Task}},
        \mathbf{F}^{\mathrm{VLM}},
        \mathbf{F}^{\mathrm{VLM}}
      \right).
    \label{eq:task_feature}
\end{equation}
The resulting $\mathbf{F}^{\mathrm{Task}}$ is subsequently projected into the spatial distance perception branch and initializes the spatial distance query $\mathbf{Q}^{\mathrm{Spa}}_0$ as follows.
\begin{equation}
    \mathbf{Q}^{\mathrm{Spa}}_0
    = \mathbf{Q}^{\mathrm{Spa}}_{\mathrm{Init}}
    + P_{\mathrm{Spa}}\!\left(\mathbf{F}^{\mathrm{Task}}\right),
    \label{eq:spa_query_init}
\end{equation}
where $\mathbf{Q}^{\mathrm{Spa}}_{\mathrm{Init}}$ and $\mathcal{P}^{\mathrm{Spa}}(\cdot)$ denote the learnable query and projection function corresponding to the spatial distance perception branch, respectively.

Starting from $\mathbf{Q}^{\mathrm{Spa}}_0$, the spatial distance perception branch extracts a task-relevant spatial distance feature from $\mathbf{F}^{\mathrm{Spa}}$ across $M$ interaction layers.  In the $m$-th layer, the interaction process can be formulated as follows.
\begin{equation}
\scalebox{0.93}{$\displaystyle
    \widetilde{\mathbf{Q}}^{\mathrm{Spa}}_m
    = \mathbf{Q}^{\mathrm{Spa}}_{m-1}
    + \CrossAttn^{\mathrm{Spa}}_m\!\left(
        \mathbf{Q}^{\mathrm{Spa}}_{m-1},
        \mathbf{F}^{\mathrm{Spa}},
        \mathbf{F}^{\mathrm{Spa}}
      \right).
$}
\label{eq:spa_crossattn}
\end{equation}
The cross-attention operation retrieves task-relevant information from $\mathbf{F}^{\mathrm{Spa}}$, and the intermediate representation is subsequently refined by a feed-forward network (FFN) with a residual connection as follows.
\begin{equation}
    \mathbf{Q}^{\mathrm{Spa}}_m
    = \widetilde{\mathbf{Q}}^{\mathrm{Spa}}_m
    + \FFN^{\mathrm{Spa}}_m\!\left(
        \widetilde{\mathbf{Q}}^{\mathrm{Spa}}_m
      \right).
    \label{eq:spa_ffn}
\end{equation}
After $M$ interaction layers, the final task-relevant feature forms the spatial distance key-value bank $(\mathbf{K}^{\mathrm{Spa}},\mathbf{V}^{\mathrm{Spa}})$ for action prediction, with $\mathbf{K}^{\mathrm{Spa}}=\mathbf{V}^{\mathrm{Spa}}=\mathbf{Q}^{\mathrm{Spa}}_M$. The geometric structure branch follows the same procedure, yielding the geometric structure key-value bank $(\mathbf{K}^{\mathrm{Geo}},\mathbf{V}^{\mathrm{Geo}})$. 

Through the two perception branches, \method enables task-relevant spatial distance perception and geometric structure perception for precise action prediction.

\subsection{When?}
\label{sec:when}

Although spatial distance perception (\textit{where}) and geometric structure perception (\textit{how}) provide additional perception for precise action prediction, their respective contributions often need to be coordinated according to current task context and evolving action state. Accordingly, adaptive spatial-geometric guidance (\textit{when}) is introduced in \method.

Specifically, in the $l$-th action expert layer equipped with adaptive spatial-geometric guidance, the noisy action feature $\mathbf{F}^{\mathrm{Act}}_{l-1}$ is projected into an action query $\mathbf{Q}^{\mathrm{Act}}_l$ that reflects the current action state. The action query then separately attends to the spatial distance and geometric structure key-value banks $(\mathbf{K}^{\mathrm{Spa}}, \mathbf{V}^{\mathrm{Spa}})$ and $(\mathbf{K}^{\mathrm{Geo}}, \mathbf{V}^{\mathrm{Geo}})$, yielding $\mathbf{F}^{\mathrm{Spa}}_l$ and $\mathbf{F}^{\mathrm{Geo}}_l$, respectively. For the spatial distance perception branch, this operation is formulated as follows.
\begin{equation}
    \mathbf{F}^{\mathrm{Spa}}_l
    = \CrossAttn^{\mathrm{Spa}}_l\!\left(
        \mathbf{Q}^{\mathrm{Act}}_l,
        \mathbf{K}^{\mathrm{Spa}},
        \mathbf{V}^{\mathrm{Spa}}
      \right).
    \label{eq:action_retrieval}
\end{equation}

% To determine the respective contributions of spatial distance perception and geometric structure perception, the two retrieved features are then used to estimate the two contribution weights together with the action query $\mathbf{Q}^{\mathrm{Act}}_l$ and the task semantic feature $\mathbf{F}^{\mathrm{Task}}$ defined in \cref{eq:task_feature} as follows.

To coordinate the contributions of spatial distance perception and geometric structure perception, their contribution weights are jointly estimated from $\mathbf{F}^{\mathrm{Spa}}_l$, $\mathbf{F}^{\mathrm{Geo}}_l$, the action query $\mathbf{Q}^{\mathrm{Act}}_l$, and the task semantic feature $\mathbf{F}^{\mathrm{Task}}$.
\begin{equation}
\begin{split}
    [\mathbf{W}^{\mathrm{Spa}}_l,\mathbf{W}^{\mathrm{Geo}}_l]
    = \sigma\Bigl(&\MLP_l\bigl(\Concat(
      \mathbf{Q}^{\mathrm{Act}}_l,
      \mathbf{F}^{\mathrm{Spa}}_l,\\[-1mm]
      &\mathbf{F}^{\mathrm{Geo}}_l,
      \mathbf{F}^{\mathrm{Task}})\bigr)\Bigr),
\end{split}
    \label{eq:guidance_weights}
\end{equation}
where $\sigma(\cdot)$ denotes the sigmoid function. Consequently, the contribution weights are conditioned on both the current task context and the evolving action state.

Subsequently, $\mathbf{Q}^{\mathrm{Act}}_l$ queries the corresponding vision-language key-value cache $(\mathbf{K}^{\mathrm{VLM}}_l,\mathbf{V}^{\mathrm{VLM}}_l)$, thereby retaining semantic entity perception as the primary task-relevant context. Then $\mathbf{F}^{\mathrm{Spa}}_l$ and $\mathbf{F}^{\mathrm{Geo}}_l$ are integrated into the action prediction through a weighted residual update to produce $\mathbf{F}^{\mathrm{Act}}_l$ for the next action expert layer.

With the integration of adaptive spatial-geometric guidance, \method coordinates the contributions of spatial distance perception and geometric structure perception based on the current task context and evolving action state.

\subsection{Training and Inference}
\label{sec:training_inference}

\textbf{Training.}
\method comprises a VLM, a spatial distance perception branch equipped with a depth estimation model, a geometric structure perception branch equipped with a visual geometry model, and an action expert incorporating adaptive spatial-geometric guidance. During training, the pre-trained VLM, depth estimation model, and visual geometry model remain frozen, with optimization of the action expert and all new parameters introduced by \method. All trainable parameters are jointly optimized using the original conditional flow-matching objective of the base VLA model, without dedicated objectives for spatial distance or geometric structure perception.

\noindent\textbf{Inference.}
During inference, the VLM encodes the 2D visual observation and language instruction to produce the final vision-language feature and vision-language key-value cache. In parallel, the depth estimation model and visual geometry model process the same observation to extract the spatial distance and geometric structure features, respectively. Starting from Gaussian noise, the action expert iteratively generates an action chunk by interacting with the vision-language key-value cache and incorporating spatial distance and geometric structure perception from the corresponding key-value banks. This process extends the VLA perception to support precise action prediction.

% You must include your signed IEEE copyright release form when you submit your finished paper.
% We MUST have this form before your paper can be published in the proceedings.

% Please direct any questions to the production editor in charge of these proceedings at the IEEE Computer Society Press:
% \url{https://www.computer.org/about/contact}.

%% file: sec/4_experiments.tex
\section{Experiments}
\label{sec:experiments}

\input{tables/table_libero}

In this section, we present the experimental setup in \cref{sec:experimental_setup}, simulation results in \cref{sec:simulation_results}, ablation studies in \cref{sec:ablation}, real-world experiments in \cref{sec:real_world_results}, and qualitative analyses in \cref{sec:qualitative}.

\subsection{Experimental Setup}
\label{sec:experimental_setup}

\textbf{Simulation benchmarks.}
\method is evaluated across 4 benchmarks following the respective evaluation protocols.

\begin{itemize}[leftmargin=*, itemsep=2pt, topsep=0pt, parsep=0pt]
    \item \textbf{LIBERO.}
    LIBERO~\cite{liu2023libero} comprises 4 task suites, \textit{i.e.}, LIBERO-Spatial (Spatial), LIBERO-Object (Object), LIBERO-Goal (Goal), and LIBERO-Long (Long), each containing 10 manipulation tasks. Following the standard LIBERO protocol, the policy is trained on 50 human-teleoperated demonstrations and evaluated over 50 trials per task, totaling 2,000 evaluation trials across the 4 suites. The results are reported as the task success rate for each suite and the average across all 4 suites.

    \item \textbf{LIBERO-Plus.}
    LIBERO-Plus~\cite{fei2025libero} extends LIBERO to 10,030 evaluation episodes for evaluating policy robustness under 7 perturbation types, \textit{i.e.}, camera viewpoint (Camera), robot initial state (Robot), language instruction (Language), lighting condition (Light), background texture (Background), sensor noise (Noise), and object layout (Layout). For evaluation, the policy trained on LIBERO is directly tested on LIBERO-Plus. The results are reported as the task success rate for each type of perturbation and the overall across all 7 dimensions.

    \item \textbf{CALVIN.}
    CALVIN~\cite{mees2022calvin} comprises 34 language-guided manipulation tasks across 4 diverse environments, A, B, C, and D. Following the ABC$\rightarrow$D protocol, the policy is trained on A, B, and C and evaluated on D. The policy is evaluated on 1,000 task sequences, each comprising 5 consecutive manipulation tasks. The results are reported as the average length of consecutively completed tasks per sequence. The task success rates at each sequence length are provided in the appendix.
    
    \item \textbf{LIBERO-PRO.}
    LIBERO-PRO~\cite{zhou2025libero} extends 4 LIBERO task suites with perturbations, \textit{i.e.}, object attributes, initial spatial configurations, semantic instruction variations, and task goals. Following the standard LIBERO-PRO evaluation protocol, the policy trained on LIBERO is directly tested on LIBERO-PRO. The overall task success rate is reported, with detailed results provided in the appendix.
\end{itemize}

\input{tables/table_libero_plus}

\noindent\textbf{Real-world experiments.}
We further evaluate \method through real-world experiments on the AgileX Piper robot. Specifically, 4 tasks are designed to evaluate semantic entity perception (\textit{which}), spatial distance perception (\textit{where}), geometric structure perception (\textit{how}), and adaptive spatial-geometric guidance (\textit{when}), respectively. For each task, policies are trained on 50 human-teleoperated demonstrations and evaluated on 20 trials. A trial is successful only if the full manipulation is completed and the task-specific goal is achieved. The success rate for each task and the average across all 4 tasks are reported.

\begin{itemize}[leftmargin=*, itemsep=2pt, topsep=0pt, parsep=0pt]
    \item \textbf{Which.}
    Given various geometric blocks, the robot is instructed to pick and place the specified one on the target plate. During testing, the block positions are varied across trials, thereby evaluating semantic entity perception for identifying \textit{which} entities are relevant to action prediction.

    \item \textbf{Where.}
    Given 2 cubes at different distances from the robot and a stepped platform, the robot is instructed to place the nearer or farther cube on the higher or lower step. The initial cube positions and target placement heights are varied during testing, thereby evaluating spatial distance perception for determining \textit{where} relevant entities are located within the spatial layout.

    \item \textbf{How.}
    Given a cube and a shape-sorting box, the robot is instructed to grasp, align, and insert the cube into the matching opening. The initial orientations of the cube and the target opening are varied during testing, thereby evaluating geometric structure perception to determine \textit{how} geometric details constrain physical interaction.
    
    \item \textbf{When.}
    Given 3 cubes, the robot is instructed to stack them into a tower. During stacking, spatial distance perception supports adaptation to increasing placement heights, and geometric structure perception guides precise alignment and placement. The cube sizes are varied during testing. This task is designed to evaluate adaptive spatial-geometric guidance in determining \textit{when} spatial distance or geometric structure perception should make a greater contribution to precise action prediction.
\end{itemize}

\noindent\textbf{Implementation details.}
Visual observation $\mathbf{I}_t$ comprises one environment camera view and one wrist camera view, with each resized to $224\!\times\!224$ for the VLM, $518\!\times\!518$ for the depth estimation model, and $512\!\times\!512$ for the visual geometry model. The action horizon is set to $H=10$ in simulation and $H=30$ for real-world deployment. The VLM comprises $N=18$ layers with hidden dimension $d_{\mathrm{VLM}}=2048$, and the action expert has 18 layers with $d_{\mathrm{Act}}=1024$. The spatial distance and geometric structure perception branches each employ $M=3$ interaction layers for task-relevant perception feature extraction. Adaptive spatial-geometric guidance is incorporated into 6 action expert layers at regular intervals. The depth estimation and visual geometry models are initialized with pre-trained Depth Anything V2-L \cite{yang2024depth} and VGGT-$\Omega$ \cite{wang2026vggt}, respectively.

\subsection{Simulation Results}
\label{sec:simulation_results}

\textbf{LIBERO.}
As shown in \Cref{tab:libero}, \method achieves the highest average success rate of 98.9\% across the 4 suites. Compared with the base model, \method achieves consistent improvements across all 4 suites, with the largest gain of +5.2\% on Long and the average improvement of +2.0\%. The notable 99.6\% success rate on both Spatial and Object reflects reliable action prediction across diverse spatial relationships and object types, suggesting that spatial distance perception and geometric structure perception extend semantic entity perception for precise manipulation.

\input{tables/table_additional_benchmarks}

\noindent\textbf{LIBERO-Plus.}
\method achieves an overall success rate of 88.2\% across 7 perturbation types in \Cref{tab:libero_plus}. Specifically, \method outperforms the base model across all 7 perturbation types, with the overall gain of +2.5\% and the largest gain of +5.3\% on Language. Improvements of +3.3\%, +2.2\%, and +1.5\% are achieved on Camera, Robot, and Layout, respectively. These gains suggest that \method maintains task semantic understanding while using spatial distance and geometric structure information to support reliable action prediction under perturbations.

\noindent\textbf{CALVIN.}
As shown in \Cref{tab:additional_benchmarks}, \method achieves the best average completed length of 4.46 on CALVIN, outperforming the base model by +0.15. The improvement can be attributed to spatial distance and geometric structure perception, with adaptive spatial-geometric guidance dynamically coordinating their contributions for reliable action prediction during long-horizon manipulation.

\noindent\textbf{LIBERO-PRO.}
As reported in \Cref{tab:additional_benchmarks}, \method achieves an overall success rate of 62.2\%, exceeding the base model by 8.9\%. The gain under diverse perturbations highlights the robustness of \method, suggesting that adaptively coordinating spatial distance and geometric structure perception can enrich the perceptual information available for reliable action prediction.

\subsection{Ablation Study}
\label{sec:ablation}

To assess the contributions of spatial distance perception (\textit{where}), geometric structure perception (\textit{how}), and adaptive spatial-geometric guidance (\textit{when}), the ablation studies are conducted on the LIBERO benchmark, with the results reported in \Cref{tab:ablation_components}, \Cref{tab:ablation_where_how}, and \Cref{tab:ablation_when}.

\input{tables/table_ablation_components}

\input{tables/table_ablation_where_how}

\noindent{\ensuremath{\bm{\mathcal{W}^{3}\mathcal{H}}}\textbf{-VLA.}}  As shown in \Cref{tab:ablation_components}, the base model is progressively extended with spatial distance perception (\textit{where}), geometric structure perception (\textit{how}), and adaptive spatial-geometric guidance (\textit{when}). Compared with baseline \#A1, methods \#A2 and \#A3 respectively introduce spatial distance perception and geometric structure perception by directly using their corresponding key-value banks to guide action prediction, achieving the average success rate improvement of 0.9\% and 1.0\%, respectively. The method of \#A4 further raises the average success rate to 98.3\% by combining the two forms of perception, confirming their complementary contributions of capturing spatial relationships and reasoning about geometric interaction constraints for action prediction. By further introducing the adaptive spatial-geometric guidance, the complete \method of \#A5 achieves the average success rate of 98.9\% and outperforms the baseline across all 4 suites, with the largest gain of +5.2\% on the Long suite.

% As shown in \Cref{tab:ablation_components}, the complete \method in \#A1 achieves an average success rate of 98.9\%, with adaptive spatial-geometric guidance (\textit{when}) incorporated into 6 action expert layers. Reducing the number of layers to 3 in \#A2 or increasing it to 9 in \#A3 lowers the average success rate to 98.5\% and 98.7\%, respectively, supporting the effectiveness of our design for integration with the action expert and suggesting that too many guidance interactions increase complexity without improving action prediction. In \#A4, spatial distance and geometric structure perception is directly incorporated into the corresponding action expert layers, yielding a lower average success rate of 98.3\% and highlighting the benefit of adaptive perceptual coordination. Further removing spatial distance perception (\textit{where}) in \#A5 or geometric structure perception (\textit{how}) in \#A6 reduces the average success rate to 98.0\% and 97.8\%, respectively. These results support the complementary contributions of capturing spatial relationships and geometric interaction constraints for precise action prediction.

\input{tables/table_ablation_when}

\begin{figure}[t]
    \centering
    \includegraphics[width=0.95\columnwidth]{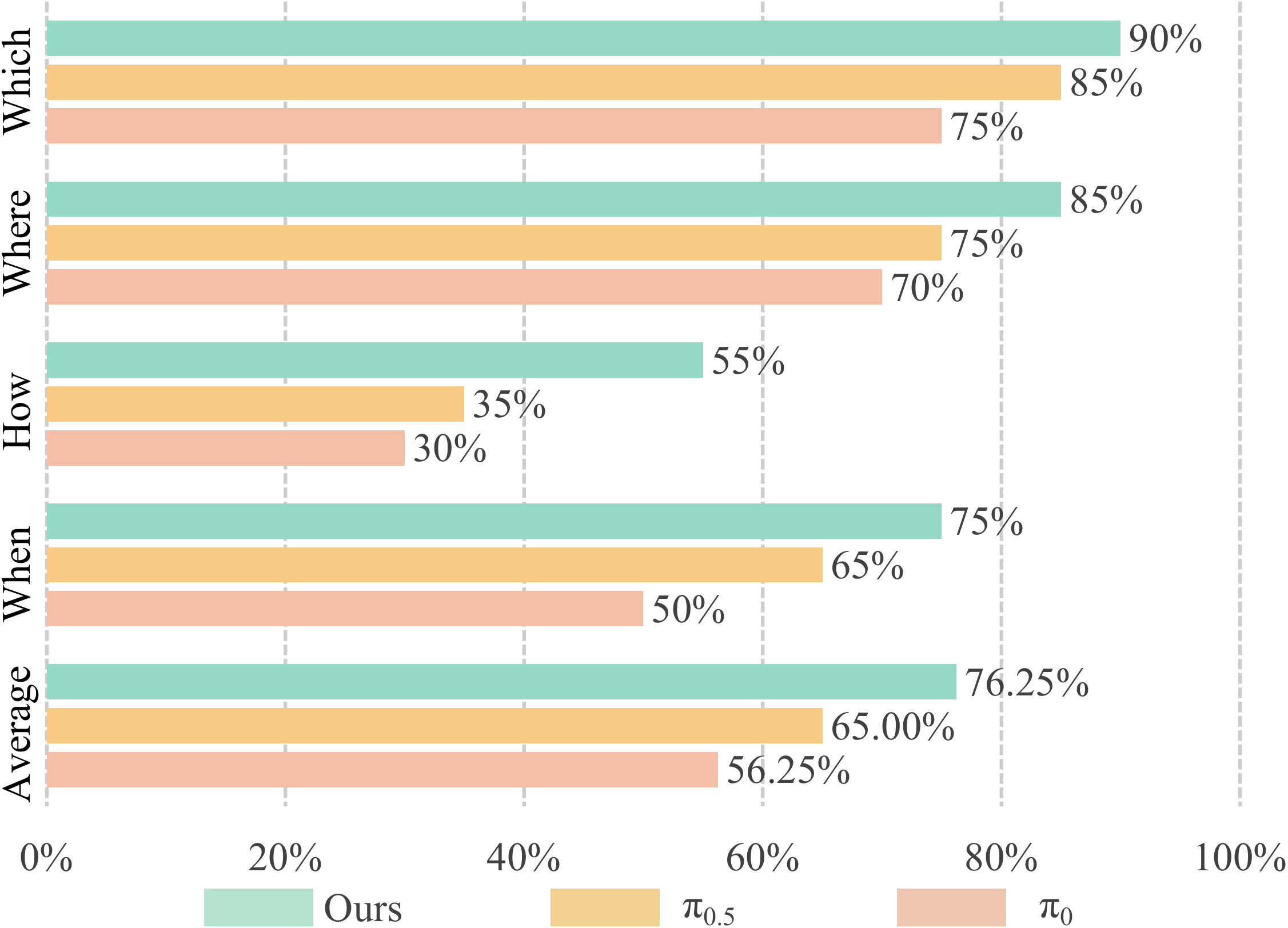}
    \caption{Comparison on the real-world experiments.}
    \label{fig:real-world}
\end{figure}

\noindent\textbf{\textit{Where} and \textit{How}.}
In \Cref{tab:ablation_where_how}, we analyze the design of spatial distance perception (\textit{where}) and geometric structure perception (\textit{how}) along two dimensions, whether the two forms of perception are organized into separate key-value banks and whether task semantics is involved in the task-relevant perception feature extraction. Compared with \#B1, \#B2 incorporates task semantics into the extraction of spatial distance and geometric structure information, increasing the average success rate to 97.5\%. About \#B3, separate key-value banks for the two forms of perception without task semantics yield a higher average success rate of 98.1\%. The complete design in \#B4 further improves the average success rate to 98.3\%. These results demonstrate the complementary benefits of separate key-value banks and task-semantic guidance for reliable action prediction.

\noindent\textbf{\textit{When}.}
As shown in \Cref{tab:ablation_when}, the design of adaptive spatial-geometric guidance is analyzed by evaluating the benefits of incorporating the current task semantic context and evolving action state into perceptual coordination. Compared with \#C1, \#C2 and \#C3 improve the average success rate to 98.4\% and 98.6\% by incorporating task semantics and action state, respectively. The complete \method in \#C4 integrates both factors, yielding the highest average success rate of 98.9\%. Task semantic context captures task requirements, and the evolving action state reflects the current execution progress, enabling \method to determine when the spatial distance or geometric structure perception should make a greater contribution to precise action prediction.

\begin{figure}[t]
    \centering
    \includegraphics[width=\columnwidth]{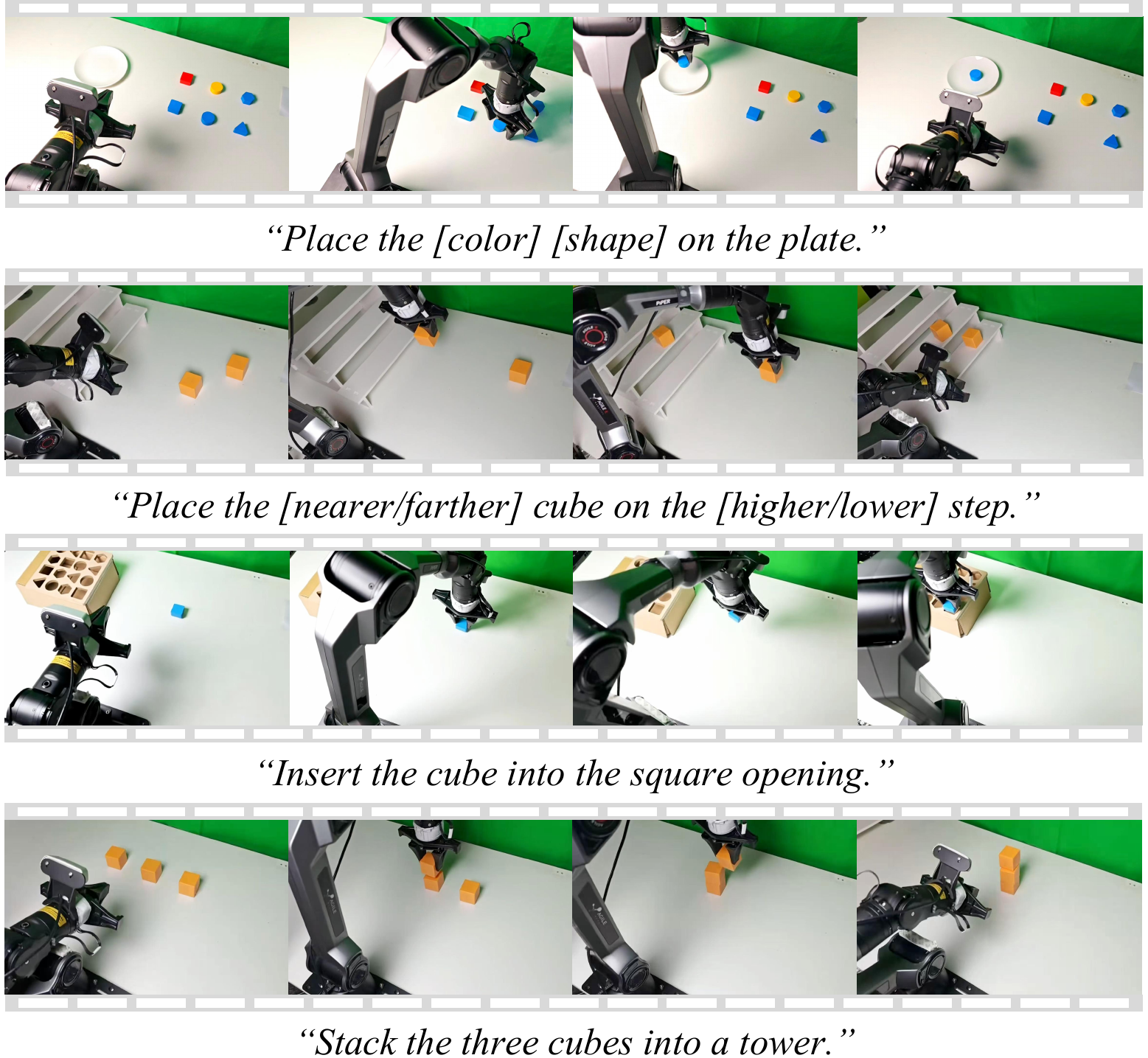}
    \caption{Visualization of the real-world manipulation tasks.}
    \label{fig:real-visualization}
\end{figure}

\subsection{Real-World Experiments}
\label{sec:real_world_results}
As shown in \cref{fig:real-world}, \method achieves an average success rate of 76.25\% across the 4 real-world tasks, surpassing \piHalf and \piZero by 11.25\% and 20\%, respectively. For \textit{Which}, \method achieves a success rate of 90\%, exceeding the base model by 5\%. Successful manipulation of the specified block among distractors indicates that \method maintains effective semantic entity perception for task-relevant entity identification. About \textit{Where}, \method achieves the best result of 85\%, supporting the benefit of spatial distance perception in capturing spatial relationships information for action prediction. The largest gain is observed on \textit{How}, where \method achieves a success rate of 55\%, compared with 35\% for the base model. Although insertion is challenging, the improvement highlights the effectiveness of geometric structure perception in capturing task-relevant geometric structure information for precise action prediction. Finally, \method achieves a success rate of 75\% on \textit{When}, exceeding the base model by 10\%. Stacking cubes requires adaptation to changing placement heights and maintaining precise alignment and stable placement, highlighting the benefit of adaptive spatial-geometric guidance. Visualization is illustrated in \cref{fig:real-visualization}.

\begin{figure}[t]
    \centering
    \includegraphics[width=0.93\columnwidth]{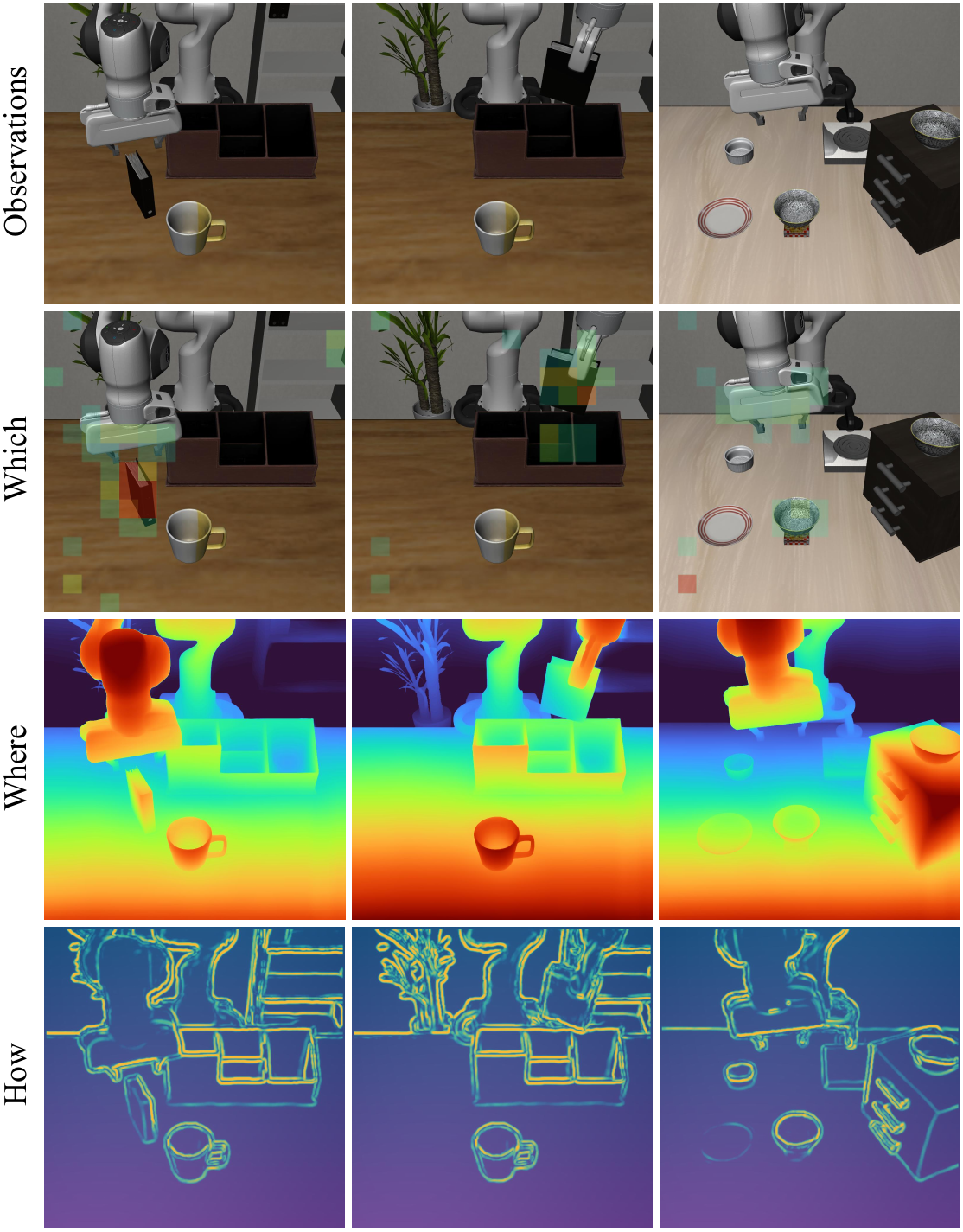}
    \caption{Visualization of perceptual inputs (Visual observations, action-visual attention maps (\textit{which}), spatial distance response maps (\textit{where}), and geometric structure response maps (\textit{how})).}
    \label{fig:qualitative}
\end{figure}

\subsection{Qualitative Analysis}
\label{sec:qualitative}
To qualitatively evaluate the effectiveness of \method, \cref{fig:qualitative} presents action-visual attention maps (\textit{which}), spatial distance response maps (\textit{where}), and geometric structure response maps (\textit{how}). For \textit{which}, the action-visual attention maps highlight the robot end effector, task-relevant entities, and interaction regions, indicating effective semantic entity perception. For \textit{where}, the spatial distance response maps capture relative spatial relationship information, providing spatial distance perception for action prediction. For \textit{how}, the geometric structure response maps emphasize the shapes and local geometric details, providing geometric structure perception for precise manipulation.

%% file: tables/table_libero.tex
\begin{table*}[t]
\centering
\caption{Comparison on the LIBERO benchmark. The best results are shown in \textbf{bold}, and the second best results are \underline{underlined}. Avg. denotes the mean of the success rates for the 4 task suites.}
\label{tab:libero}
\small
\setlength{\tabcolsep}{13pt}
\begin{tabular}{
@{}
>{\columncolor{white}[0pt][\tabcolsep]}l
|c|c|c|c|c|c
}
\toprule
\textbf{Method} & \textbf{Publication} & \textbf{Spatial} & \textbf{Object} & \textbf{Goal} & \textbf{Long} & \textbf{Avg.} \\
\midrule
% Diffusion Policy & RSS 2023 & 78.3 & 92.5 & 68.3 & 50.5 & 72.4 \\
Octo \cite{team2024octo} & RSS 2024 & 78.9 & 85.7 & 84.6 & 51.1 & 75.1 \\
OpenVLA \cite{kim2024openvla} & CoRL 2024 & 84.7 & 88.4 & 79.2 & 53.7 & 76.5 \\
TraceVLA \cite{zheng2025tracevla} & ICLR 2025 & 84.6 & 85.2 & 75.1 & 54.1 & 74.8 \\
CoT-VLA \cite{zhao2025cot} & CVPR 2025 & 87.5 & 91.6 & 87.6 & 69.0 & 83.9 \\
SpatialVLA \cite{qu2025spatialvla} & RSS 2025 & 88.2 & 89.9 & 78.6 & 55.5 & 78.1 \\
4D-VLA \cite{zhang20264d} & NeurIPS 2025 & 88.9 & 95.2 & 90.9 & 79.1 & 88.6 \\
\piZero-FAST \cite{pertsch2025fast} & RSS 2025 & 96.4 & 96.8 & 88.6 & 60.2 & 85.5 \\
\piZero \cite{black2024pi_0} & RSS 2025 & 96.8 & 98.8 & 95.8 & 85.2 & 94.2 \\
UniVLA \cite{bu2025univla} & RSS 2025 & 96.5 & 96.8 & 95.6 & 92.0 & 95.2 \\
OpenVLA-OFT \cite{kim2025fine} & RSS 2025 & 97.6 & 98.4 & 97.9 & 94.5 & 97.1 \\
\piHalf \cite{intelligence2025pi_} & CoRL 2025 & \second{98.8} & 98.2 & 98.0 & 92.4 & 96.9 \\
VLA-Adapter \cite{wang2026vla} & AAAI 2026 & 97.8 & 99.2 & 97.2 & 95.0 & 97.3 \\
MemoryVLA \cite{shi2026memoryvla} & ICLR 2026 & 98.4 & 98.4 & 96.4 & 93.4 & 96.7 \\
X-VLA \cite{zheng2026x} & ICLR 2026 & 98.2 & 98.6 & 97.8 & \best{97.6} & 98.1 \\
VIPA-VLA \cite{feng2026spatial} & CVPR 2026 & 96.6 & 98.6 & 97.0 & 95.0 & 96.8 \\
GeoPredict \cite{qian2026geopredict} & CVPR 2026 & 98.0 & 98.2 & 95.7 & 94.0 & 96.5 \\
ConsisVLA-4D \cite{li2026consisvla} & CVPR 2026 & \second{98.8} & \best{99.8} & 98.0 & 95.6 & 98.1 \\
ACoT-VLA \cite{zhong2026acot} & CVPR 2026 & 98.6 & 99.0 & \best{99.4} & \second{97.0} & \second{98.5} \\
% ACoT-VLA$^{*}$ & CVPR 2026 & \best{99.4} & \second{99.6} & 98.8 & 96.0 & \second{98.5} \\
\midrule
\rowcolor{oursbg}
\textbf{\method} & Ours & \best{99.6} & \second{99.6} & \second{98.8} & \best{97.6} & \best{98.9} \\
\bottomrule
\end{tabular}
\end{table*}

%% file: tables/table_libero_plus.tex
\begin{table*}[t]
\centering
\caption{Comparison on the LIBERO-Plus benchmark. The best results are shown in \textbf{bold}, and the second best results are \underline{underlined}.}
\label{tab:libero_plus}
\small
% \scriptsize
\setlength{\tabcolsep}{6pt}
\begin{tabular}{
@{}
>{\columncolor{white}[0pt][\tabcolsep]}l
|c|c|c|c|c|c|c|c|c
}
\toprule
\textbf{Method} & \textbf{Publication} & \textbf{Camera} & \textbf{Robot} & \textbf{Language} & \textbf{Light} & \textbf{Background} & \textbf{Noise} & \textbf{Layout} & \textbf{Overall} \\
\midrule

OpenVLA \cite{kim2024openvla} & CoRL 2024
& 0.8 & 3.5 & 23.0 & 8.1 & 34.8 & 15.2 & 28.5 & 15.6 \\

UniVLA \cite{bu2025univla} & RSS 2025
& 1.8 & 46.2 & 69.6 & 69.0 & 81.0 & 21.2 & 31.9 & 42.9 \\

\piZero-FAST \cite{pertsch2025fast} & RSS 2025
& 65.1 & 21.6 & 61.0 & 73.2 & 73.2 & 74.4 & 68.8 & 61.6 \\

OpenVLA-OFT \cite{kim2025fine} & RSS 2025
& 56.4 & 31.9 & 79.5 & 88.7 & 93.3 & 75.8 & 74.2 & 69.6 \\

\piZero \cite{black2024pi_0} & RSS 2025
& 61.0 & 40.8 & 63.5 & 89.3 & 84.1 & 80.1 & 76.4 & 69.4 \\

\piHalf \cite{intelligence2025pi_} & CoRL 2025
& \second{75.8} & 79.4 & 83.3 & 95.5
& 95.0 & \second{89.6} & 87.0 & 85.7 \\

DreamVLA \cite{zhang2026dreamvla} & NeurIPS 2025
& 65.0 & 40.9 & 63.5 & 85.7 & 82.7 & 85.0 & 74.0 & 69.9 \\

GuidedVLA \cite{jia2026guidedvla} & RSS 2026
& 73.7 & 51.4 & 62.6 & 94.6 & 89.0 & 85.2 & 79.9 & 75.4 \\

Libra-VLA \cite{wei2026libra} & ACL 2026
& 68.9 & 48.8 & \best{92.7} & \best{97.9}
& 93.4 & 86.3 & 77.5 & 79.5 \\

VLA-Adapter \cite{wang2026vla} & AAAI 2026
& 36.2 & 37.9 & 74.6 & 70.6 & 76.1 & 58.0 & 69.7 & 59.1 \\

AVA-VLA \cite{xiao2026ava} & CVPR 2026
& 55.5 & 25.9 & 85.6 & 95.5 & 88.9 & 78.0 & 74.1 & 70.1 \\

% AVA-VLA$^{\dagger}$ & CVPR 2026
% & 69.4 & 34.9 & 81.5 & 97.5 & 94.1 & 79.1 & 78.3 & 74.7 \\

ACoT-VLA \cite{zhong2026acot} & CVPR 2026
& 72.6 & \best{82.6} & 87.5 & \second{97.7}
& \second{96.5} & 87.8 & \second{88.1} & \second{86.6} \\

ResVLA \cite{zhong2026noise} & ICML 2026
& 53.2 & 57.5 & 88.2 & 94.5 & 96.3 & 81.8 & 78.3 & 76.9 \\

% LaMP & ECCV 2026
% & 64.5 & 69.6 & 88.2 & 95.3 & \best{97.4} & 76.9 & 73.8 & 79.3 \\

% ACoT-VLA$^{*}$ & CVPR 2026
% & 68.9 & 80.3 & 84.1 & 95.6
% & 93.1 & 81.5 & \second{88.3} & 83.6 \\

\midrule
\rowcolor{oursbg}
\textbf{\method} & Ours
& \best{79.1}
& \second{81.6}
& \second{88.6}
& 96.6
& \best{97.9}
& \best{90.2}
& \best{88.5}
& \best{88.2} \\
\bottomrule
\end{tabular}
\end{table*}

%% file: tables/table_additional_benchmarks.tex
% \begin{table}[t]
% \centering
% \caption{
% Comparison on the CALVIN and LIBERO-PRO benchmarks.
% The best results are shown in \textbf{bold}, and the second-best
% results are \underline{underlined}.}
% \label{tab:additional_benchmarks}
% \small
% \setlength{\tabcolsep}{6pt}

% \begin{tabular}{l|c|c|c}
% \toprule
% \multirow{2}{*}{\textbf{Method}}
% & \multirow{2}{*}{\textbf{Publication}}
% & \textbf{CALVIN}
% & \textbf{LIBERO-PRO} \\
% & & \textbf{Len.} & \textbf{SR} \\
% \midrule
% OpenVLA
% & CoRL 2024
% & 3.27
% & 52.0 \\
% $\pi_{0}$
% & RSS 2025
% & 3.92
% & 44.0 \\
% $\pi_{0.5}$
% & CoRL 2025
% & 4.26
% & \second{53.0} \\
% X-VLA
% & ICLR 2026
% & \second{4.43}
% & 46.0 \\
% \midrule
% \rowcolor{oursbg}
% \textbf{\method}
% & Ours
% & \best{4.46}
% & \best{61.9} \\
% \bottomrule
% \end{tabular}
% \end{table}

\begin{table}[t]
\centering
\caption{
Comparison on the CALVIN and LIBERO-PRO benchmarks.
The best results are shown in \textbf{bold}, and the second best results are \underline{underlined}.}
\label{tab:additional_benchmarks}
\small
\setlength{\tabcolsep}{2.8pt}

\begin{tabular}{l|c|c|c}
\toprule
\multirow{2}{*}{\textbf{Method}}
& \multirow{2}{*}{\textbf{Publication}}
& \textbf{CALVIN}
& \textbf{LIBERO-PRO} \\
& & \textbf{Length} & \textbf{Success Rate} \\
\midrule
OpenVLA \cite{kim2024openvla}
& CoRL 2024
& 3.27
& 47.6 \\
$\pi_{0}$ \cite{black2024pi_0}
& RSS 2025
& 3.84
& 45.3 \\
OpenVLA-OFT \cite{kim2025fine}
& RSS 2025
& 4.10
& 23.3 \\
$\pi_{0.5}$ \cite{intelligence2025pi_}
& CoRL 2025
& 4.31
& \second{53.3} \\
VLA-Adapter \cite{wang2026vla}
& AAAI 2026
& 4.42
& 46.1 \\
X-VLA \cite{zheng2026x}
& ICLR 2026
& \second{4.43}
& 44.8 \\
\midrule
\rowcolor{oursbg}
\textbf{\method}
& Ours
& \best{4.46}
& \best{62.2} \\
\bottomrule
\end{tabular}
\end{table}

%% file: tables/table_ablation_components.tex
\begin{table}[t]
\centering
\caption{Ablation of \method on the LIBERO benchmark.}
\label{tab:ablation_components}
\small
\setlength{\tabcolsep}{4.3pt}
\begin{tabular}{@{}c|l|cccc|c@{}}
\toprule
\textbf{ID}
& \textbf{Method}
& \textbf{Spatial}
& \textbf{Object}
& \textbf{Goal}
& \textbf{Long}
& \textbf{Avg.} \\
\midrule
\#A1 & Baseline      & 98.8 & 98.2 & 98.0 & 92.4 & 96.9 \\
\#A2 & + Where       & 99.0 & 98.6 & 98.0 & 95.6 & 97.8 \\
\#A3 & + How         & 99.4 & 98.6 & 98.4 & 95.2 & 97.9 \\
\#A4 & + Where + How & 99.4 & 98.8 & 98.2 & 96.6 & 98.3 \\
\#A5 & \textbf{\method}
                    & \textbf{99.6}
                    & \textbf{99.6}
                    & \textbf{98.8}
                    & \textbf{97.6}
                    & \textbf{98.9} \\
\bottomrule
\end{tabular}
\end{table}
% \begin{table}[t]
% \centering
% \caption{Ablation of \method on the LIBERO benchmark.}
% \label{tab:ablation_components}
% \small
% \setlength{\tabcolsep}{3.2pt}
% \begin{tabular}{@{}c|c|cccc|c@{}}
% \toprule
% \textbf{ID}
% & \textbf{Method}
% & \textbf{Spatial}
% & \textbf{Object}
% & \textbf{Goal}
% & \textbf{Long}
% & \textbf{Avg.} \\
% \midrule
% \#A1 & \textbf{\method}
% & \textbf{99.6}
% & \textbf{99.6}
% & \textbf{98.8}
% & \textbf{97.6}
% & \textbf{98.9} \\
% \#A2 & w/ When (3 layers)
% & 99.4 & 99.4 & 98.4 & 96.8 & 98.5 \\
% \#A3 & w/ When (9 layers)
% & 99.6 & 99.2 & 98.6 & 97.2 & 98.7 \\
% \#A4 & w/o When
% & 99.4 & 98.8 & 98.2 & 96.6 & 98.3 \\
% \#A5 & w/o When \& Where
% & 99.4 & 98.6 & 98.4 & 95.2 & 98.0 \\
% \#A6 & w/o When \& How
% & 99.0 & 98.6 & 98.0 & 95.6 & 97.8 \\
% \bottomrule
% \end{tabular}
% \end{table}

%% file: tables/table_ablation_where_how.tex
\begin{table}[t]
\centering
\caption{
Ablation of spatial distance perception (\textit{where}) and
geometric structure perception (\textit{how}) on the LIBERO benchmark.}
\label{tab:ablation_where_how}
\small
\setlength{\tabcolsep}{1.5pt}

\begin{tabular}{@{}c|cc|cccc|c@{}}
\toprule
\multirow{2}{*}{\textbf{ID}}
& \textbf{Separate}
& \textbf{Task}
& \multirow{2}{*}{\textbf{Spatial}}
& \multirow{2}{*}{\textbf{Object}}
& \multirow{2}{*}{\textbf{Goal}}
& \multirow{2}{*}{\textbf{Long}}
& \multirow{2}{*}{\textbf{Avg.}} \\
& \textbf{Organization}
& \textbf{Semantics}
& & & & & \\
\midrule
\#B1
& \xmark
& \xmark
& 98.8
& 98.4
& 97.8
& 93.6
& 97.2 \\
\#B2
& \xmark
& \cmark
& 99.0
& 98.6
& 98.0
& 94.2
& 97.5 \\
\#B3
& \cmark
& \xmark
& 99.2
& \textbf{99.0}
& \textbf{98.2}
& 96.0
& 98.1 \\
\#B4
& \cmark
& \cmark
& \textbf{99.4}
& 98.8
& \textbf{98.2}
& \textbf{96.6}
& \textbf{98.3} \\
\bottomrule
\end{tabular}
\end{table}

%% file: tables/table_ablation_when.tex
\begin{table}[t]
\centering
\caption{
Ablation of adaptive spatial-geometric guidance (\textit{when})
on the LIBERO benchmark.}
\label{tab:ablation_when}
\small
\setlength{\tabcolsep}{2.8pt}

\begin{tabular}{@{}c|cc|cccc|c@{}}
\toprule
\multirow{2}{*}{\textbf{ID}}
& \textbf{Action}
& \textbf{Task}
& \multirow{2}{*}{\textbf{Spatial}}
& \multirow{2}{*}{\textbf{Object}}
& \multirow{2}{*}{\textbf{Goal}}
& \multirow{2}{*}{\textbf{Long}}
& \multirow{2}{*}{\textbf{Avg.}} \\
& \textbf{State}
& \textbf{Semantics}
& & & & & \\
\midrule
\#C1
& --
& --
& 99.4
& 98.8
& 98.2
& 96.6
& 98.3 \\
\#C2
& \xmark
& \cmark
& 99.2
& 99.0
& \textbf{98.8}
& 96.4
& 98.4 \\
\#C3
& \cmark
& \xmark
& \textbf{99.6}
& 99.4
& 98.4
& 97.0
& 98.6 \\
\#C4
& \cmark
& \cmark
& \textbf{99.6}
& \textbf{99.6}
& \textbf{98.8}
& \textbf{97.6}
& \textbf{98.9} \\
\bottomrule
\end{tabular}
\end{table}

%% file: sec/5_conclusion.tex
\section{Conclusion}
\label{sec:conclusion} 
In this work, we address the perceptual information requirements for precise action prediction in VLA models by introducing \method, a framework that organizes perception and action prediction around \textit{which}, \textit{where}, \textit{how}, and \textit{when}. Specifically, \method retains VLM-based semantic entity perception and introduces spatial distance and geometric structure perception to capture spatial relationships information and geometric interaction constraints. Adaptive spatial-geometric guidance further coordinates their contributions according to the task semantic context and evolving action state. Together, these designs enrich VLA perception with spatial and geometric information for precise action prediction in the complex physical world.

%% file: sec/6_llm.tex
\section*{LLM Usage Statement}
OpenAI ChatGPT assisted with manuscript writing, searches for related work, code debugging, experimental design, figure layout, and data organization.

%% file: main.bib
@String(CVPR= {IEEE Conf. Comput. Vis. Pattern Recog.})

@String(AAAI = {AAAI})

@String(CVPR  = {CVPR})

@article{team2024octo,
  title={Octo: An open-source generalist robot policy},
  author={Team, Octo Model and Ghosh, Dibya and Walke, Homer and Pertsch, Karl and Black, Kevin and Mees, Oier and Dasari, Sudeep and Hejna, Joey and Kreiman, Tobias and Xu, Charles and others},
  journal={arXiv preprint arXiv:2405.12213},
  year={2024}
}

@article{achiam2023gpt,
  title={GPT-4 technical report},
  author={Achiam, J},
  journal={arXiv preprint arXiv:2303.08774},
  year={2023}
}

@article{brohan2022rt,
  title={Rt-1: Robotics transformer for real-world control at scale},
  author={Brohan, Anthony and Brown, Noah and Carbajal, Justice and Chebotar, Yevgen and Dabis, Joseph and Finn, Chelsea and Gopalakrishnan, Keerthana and Hausman, Karol and Herzog, Alex and Hsu, Jasmine and others},
  journal={arXiv preprint arXiv:2212.06817},
  year={2022}
}

@inproceedings{zitkovich2023rt,
  title={Rt-2: Vision-language-action models transfer web knowledge to robotic control},
  author={Zitkovich, Brianna and Yu, Tianhe and Xu, Sichun and Xu, Peng and Xiao, Ted and Xia, Fei and Wu, Jialin and Wohlhart, Paul and Welker, Stefan and Wahid, Ayzaan and others},
  booktitle={Conference on robot learning},
  pages={2165--2183},
  year={2023},
  organization={PMLR}
}

@article{kim2024openvla,
  title={Openvla: An open-source vision-language-action model},
  author={Kim, Moo Jin and Pertsch, Karl and Karamcheti, Siddharth and Xiao, Ted and Balakrishna, Ashwin and Nair, Suraj and Rafailov, Rafael and Foster, Ethan and Lam, Grace and Sanketi, Pannag and others},
  journal={arXiv preprint arXiv:2406.09246},
  year={2024}
}

@article{black2024pi_0,
  title={{$\pi_0$}: A Vision-Language-Action Flow Model for General Robot Control},
  author={Black, Kevin and Brown, Noah and Driess, Danny and Esmail, Adnan and Equi, Michael and Finn, Chelsea and Fusai, Niccolo and Groom, Lachy and Hausman, Karol and Ichter, Brian and others},
  journal={arXiv preprint arXiv:2410.24164},
  year={2024}
}

@article{intelligence2025pi_,
  title={{$\pi_{0.5}$}: a Vision-Language-Action Model with Open-World Generalization},
  author={Intelligence, Physical and Black, Kevin and Brown, Noah and Darpinian, James and Dhabalia, Karan and Driess, Danny and Esmail, Adnan and Equi, Michael and Finn, Chelsea and Fusai, Niccolo and others},
  journal={arXiv preprint arXiv:2504.16054},
  year={2025}
}

@inproceedings{zheng2026x,
  title={X-vla: Soft-prompted transformer as scalable cross-embodiment vision-language-action model},
  author={Zheng, Jinliang and Li, Jianxiong and Wang, Zhihao and Liu, Dongxiu and Kang, Xirui and Feng, Yuchun and Zheng, Yinan and Zou, Jiayin and Chen, Yilun and Zeng, Jia and others},
  booktitle={International Conference on Learning Representations},
  volume={2026},
  pages={60580--60606},
  year={2026}
}

@article{luo2026simvla,
  title={Simvla: A simple vla baseline for robotic manipulation},
  author={Luo, Yuankai and Chen, Woping and Liang, Tong and Wang, Baiqiao and Li, Zhenguo},
  journal={arXiv preprint arXiv:2602.18224},
  year={2026}
}

@inproceedings{wang2026vggt,
  title={{VGGT-$\Omega$}},
  author={Wang, Jianyuan and Chen, Minghao and Zhang, Shangzhan and Karaev, Nikita and Sch{\"o}nberger, Johannes and Labatut, Patrick and Bojanowski, Piotr and Novotny, David and Vedaldi, Andrea and Rupprecht, Christian},
  booktitle={Conference on Computer Vision and Pattern Recognition 2026},
  year={2026}
}

@article{yang2024depth,
  title={Depth anything v2},
  author={Yang, Lihe and Kang, Bingyi and Huang, Zilong and Zhao, Zhen and Xu, Xiaogang and Feng, Jiashi and Zhao, Hengshuang},
  journal={Advances in neural information processing systems},
  volume={37},
  pages={21875--21911},
  year={2024}
}

@article{liu2023libero,
  title={Libero: Benchmarking knowledge transfer for lifelong robot learning},
  author={Liu, Bo and Zhu, Yifeng and Gao, Chongkai and Feng, Yihao and Liu, Qiang and Zhu, Yuke and Stone, Peter},
  journal={Advances in Neural Information Processing Systems},
  volume={36},
  pages={44776--44791},
  year={2023}
}

@article{fei2025libero,
  title={Libero-plus: In-depth robustness analysis of vision-language-action models},
  author={Fei, Senyu and Wang, Siyin and Shi, Junhao and Dai, Zihao and Cai, Jikun and Qian, Pengfang and Ji, Li and He, Xinzhe and Zhang, Shiduo and Fei, Zhaoye and others},
  journal={arXiv preprint arXiv:2510.13626},
  year={2025}
}

@article{zhou2025libero,
  title={Libero-pro: Towards robust and fair evaluation of vision-language-action models beyond memorization},
  author={Zhou, Xueyang and Xu, Yangming and Tie, Guiyao and Chen, Yongchao and Zhang, Guowen and Chu, Duanfeng and Zhou, Pan and Sun, Lichao},
  journal={arXiv preprint arXiv:2510.03827},
  year={2025}
}

@article{mees2022calvin,
  title={Calvin: A benchmark for language-conditioned policy learning for long-horizon robot manipulation tasks},
  author={Mees, Oier and Hermann, Lukas and Rosete-Beas, Erick and Burgard, Wolfram},
  journal={IEEE Robotics and Automation Letters},
  volume={7},
  number={3},
  pages={7327--7334},
  year={2022},
  publisher={IEEE}
}

@article{qu2025spatialvla,
  title={Spatialvla: Exploring spatial representations for visual-language-action model},
  author={Qu, Delin and Song, Haoming and Chen, Qizhi and Yao, Yuanqi and Ye, Xinyi and Ding, Yan and Wang, Zhigang and Gu, JiaYuan and Zhao, Bin and Wang, Dong and others},
  journal={arXiv preprint arXiv:2501.15830},
  year={2025}
}

@inproceedings{o2024open,
  title={Open x-embodiment: Robotic learning datasets and rt-x models: Open x-embodiment collaboration 0},
  author={O’Neill, Abby and Rehman, Abdul and Maddukuri, Abhiram and Gupta, Abhishek and Padalkar, Abhishek and Lee, Abraham and Pooley, Acorn and Gupta, Agrim and Mandlekar, Ajay and Jain, Ajinkya and others},
  booktitle={2024 IEEE International Conference on Robotics and Automation (ICRA)},
  pages={6892--6903},
  year={2024},
  organization={IEEE}
}

@inproceedings{liu2025rdt,
  title={Rdt-1b: a diffusion foundation model for bimanual manipulation},
  author={Liu, Songming and Wu, Lingxuan and Li, Bangguo and Tan, Hengkai and Chen, Huayu and Wang, Zhengyi and Xu, Ke and Su, Hang and Zhu, Jun},
  booktitle={International Conference on Learning Representations},
  volume={2025},
  pages={29982--30009},
  year={2025}
}

@inproceedings{shridhar2022cliport,
  title={Cliport: What and where pathways for robotic manipulation},
  author={Shridhar, Mohit and Manuelli, Lucas and Fox, Dieter},
  booktitle={Conference on robot learning},
  pages={894--906},
  year={2022},
  organization={pmlr}
}

@inproceedings{huang2025roboground,
  title={Roboground: Robotic manipulation with grounded vision-language priors},
  author={Huang, Haifeng and Chen, Xinyi and Chen, Yilun and Li, Hao and Han, Xiaoshen and Wang, Zehan and Wang, Tai and Pang, Jiangmiao and Zhao, Zhou},
  booktitle={2025 IEEE/CVF Conference on Computer Vision and Pattern Recognition (CVPR)},
  pages={22540--22550},
  year={2025},
  organization={IEEE}
}

@inproceedings{shridhar2023perceiver,
  title={Perceiver-actor: A multi-task transformer for robotic manipulation},
  author={Shridhar, Mohit and Manuelli, Lucas and Fox, Dieter},
  booktitle={Conference on Robot Learning},
  pages={785--799},
  year={2023},
  organization={PMLR}
}

@article{gervet2023act3d,
  title={Act3d: 3d feature field transformers for multi-task robotic manipulation},
  author={Gervet, Theophile and Xian, Zhou and Gkanatsios, Nikolaos and Fragkiadaki, Katerina},
  journal={arXiv preprint arXiv:2306.17817},
  year={2023}
}

@inproceedings{goyal2023rvt,
  title={Rvt: Robotic view transformer for 3d object manipulation},
  author={Goyal, Ankit and Xu, Jie and Guo, Yijie and Blukis, Valts and Chao, Yu-Wei and Fox, Dieter},
  booktitle={Conference on Robot Learning},
  pages={694--710},
  year={2023},
  organization={PMLR}
}

@article{sun2025hypertasr,
  title={Hypertasr: Hypernetwork-driven task-aware scene representations for robust manipulation},
  author={Sun, Li and Wu, Jiefeng and Chen, Feng and Liu, Ruizhe and Yang, Yanchao},
  journal={arXiv preprint arXiv:2508.18802},
  year={2025}
}

@article{liu2026activevla,
  title={ActiveVLA: Injecting active perception into vision-language-action models for precise 3D robotic manipulation},
  author={Liu, Zhenyang and Gu, Yongchong and Wang, Yikai and Xue, Xiangyang and Fu, Yanwei},
  journal={arXiv preprint arXiv:2601.08325},
  year={2026}
}

@article{beyer2024paligemma,
  title={Paligemma: A versatile 3b vlm for transfer},
  author={Beyer, Lucas and Steiner, Andreas and Pinto, Andr{\'e} Susano and Kolesnikov, Alexander and Wang, Xiao and Salz, Daniel and Neumann, Maxim and Alabdulmohsin, Ibrahim and Tschannen, Michael and Bugliarello, Emanuele and others},
  journal={arXiv preprint arXiv:2407.07726},
  year={2024}
}

@article{lipman2022flow,
  title={Flow matching for generative modeling},
  author={Lipman, Yaron and Chen, Ricky TQ and Ben-Hamu, Heli and Nickel, Maximilian and Le, Matt},
  journal={arXiv preprint arXiv:2210.02747},
  year={2022}
}

@inproceedings{zheng2025tracevla,
  title={Tracevla: Visual trace prompting enhances spatial-temporal awareness for generalist robotic policies},
  author={Zheng, Ruijie and Liang, Yongyuan and Huang, Shuaiyi and Gao, Jianfeng and Daum{\'e} III, Hal and Kolobov, Andrey and Huang, Furong and Yang, Jianwei},
  booktitle={International Conference on Learning Representations},
  volume={2025},
  pages={54277--54296},
  year={2025}
}

@inproceedings{zhao2025cot,
  title={Cot-vla: Visual chain-of-thought reasoning for vision-language-action models},
  author={Zhao, Qingqing and Lu, Yao and Kim, Moo Jin and Fu, Zipeng and Zhang, Zhuoyang and Wu, Yecheng and Li, Zhaoshuo and Ma, Qianli and Han, Song and Finn, Chelsea and others},
  booktitle={2025 IEEE/CVF Conference on Computer Vision and Pattern Recognition (CVPR)},
  pages={1702--1713},
  year={2025},
  organization={IEEE}
}

@inproceedings{zhong2026acot,
  title={Acot-vla: Action chain-of-thought for vision-language-action models},
  author={Zhong, Linqing and Liu, Yi and Wei, Yifei and Xiong, Ziyu and Liu, Si and Ren, Guanghui},
  booktitle={Proceedings of the IEEE/CVF Conference on Computer Vision and Pattern Recognition},
  pages={8152--8162},
  year={2026}
}

@article{zhang20264d,
  title={4d-vla: Spatiotemporal vision-language-action pretraining with cross-scene calibration},
  author={Zhang, Jiahui and Chen, Yurui and Xu, Yueming and Huang, Ze and Zhou, Yanpeng and Yuan, Yu-Jie and Cai, Xinyue and Huang, Guowei and Quan, Xingyue and Xu, Hang and others},
  journal={Advances in Neural Information Processing Systems},
  volume={38},
  pages={33914--33937},
  year={2026}
}

@article{pertsch2025fast,
  title={Fast: Efficient action tokenization for vision-language-action models},
  author={Pertsch, Karl and Stachowicz, Kyle and Ichter, Brian and Driess, Danny and Nair, Suraj and Vuong, Quan and Mees, Oier and Finn, Chelsea and Levine, Sergey},
  journal={arXiv preprint arXiv:2501.09747},
  year={2025}
}

@article{bu2025univla,
  title={Univla: Learning to act anywhere with task-centric latent actions},
  author={Bu, Qingwen and Yang, Yanting and Cai, Jisong and Gao, Shenyuan and Ren, Guanghui and Yao, Maoqing and Luo, Ping and Li, Hongyang},
  journal={arXiv preprint arXiv:2505.06111},
  year={2025}
}

@article{kim2025fine,
  title={Fine-tuning vision-language-action models: Optimizing speed and success},
  author={Kim, Moo Jin and Finn, Chelsea and Liang, Percy},
  journal={arXiv preprint arXiv:2502.19645},
  year={2025}
}

@inproceedings{wang2026vla,
  title={Vla-adapter: An effective paradigm for tiny-scale vision-language-action model},
  author={Wang, Yihao and Ding, Pengxiang and Li, Lingxiao and Cui, Can and Ge, Zirui and Tong, Xinyang and Song, Wenxuan and Zhao, Han and Zhao, Wei and Hou, Pengxu and others},
  booktitle={Proceedings of the AAAI conference on artificial intelligence},
  volume={40},
  number={22},
  pages={18638--18646},
  year={2026}
}

@inproceedings{shi2026memoryvla,
  title={Memoryvla: Perceptual-cognitive memory in vision-language-action models for robotic manipulation},
  author={Shi, Hao and Xie, Bin and Liu, Yingfei and Sun, Lin and Liu, Fengrong and Wang, Tiancai and Zhou, Erjin and Fan, Haoqiang and Zhang, Xiangyu and Huang, Gao},
  booktitle={International Conference on Learning Representations},
  volume={2026},
  pages={18567--18602},
  year={2026}
}

@inproceedings{qian2026geopredict,
  title={Geopredict: Leveraging predictive kinematics and 3d gaussian geometry for precise vla manipulation},
  author={Qian, Jingjing and Han, Boyao and Shi, Chen and Xiao, Lei and Yang, Long and Shi, Shaoshuai and Jiang, Li},
  booktitle={Proceedings of the IEEE/CVF Conference on Computer Vision and Pattern Recognition},
  pages={13529--13539},
  year={2026}
}

@article{zhang2026dreamvla,
  title={Dreamvla: a vision-language-action model dreamed with comprehensive world knowledge},
  author={Zhang, Wenyao and Liu, Hongsi and Qi, Zekun and Wang, Yunnan and Yu, Xinqiang and Zhang, Jiazhao and Dong, Runpei and He, Jiawei and Wang, He and Zhang, Zhizheng and others},
  journal={Advances in Neural Information Processing Systems},
  volume={38},
  pages={24195--24228},
  year={2026}
}

@article{jia2026guidedvla,
  title={Guidedvla: Specifying task-relevant factors via plug-and-play action attention specialization},
  author={Jia, Xiaosong and Yang, Bowen and Ge, Zuhao and Nie, Xian and Zhou, Yuchen and Fan, Cunxin and Li, Yufeng and Chai, Yilin and Jing, Chao and Liang, Zijian and others},
  journal={arXiv preprint arXiv:2605.12369},
  volume={1},
  year={2026}
}

@inproceedings{wei2026libra,
  title={Libra-VLA: Achieving Learning Equilibrium via Asynchronous Coarse-to-Fine Dual-System},
  author={Wei, Yifei and Zhong, Linqing and Liu, Yi and Lu, Yuxiang and He, Xindong and Yao, Maoqing and Ren, Guanghui},
  booktitle={Proceedings of the 64th Annual Meeting of the Association for Computational Linguistics (Volume 1: Long Papers)},
  pages={39799--39815},
  year={2026}
}

@inproceedings{xiao2026ava,
  title={AVA-VLA: improving vision-language-action models with active visual attention},
  author={Xiao, Lei and Li, Jifeng and Gao, Juntao and Ye, Feiyang and Jin, Yan and Qian, Jingjing and Zhang, Jing and Wu, Yong and Yu, Xiaoyuan},
  booktitle={Proceedings of the IEEE/CVF Conference on Computer Vision and Pattern Recognition},
  pages={13453--13463},
  year={2026}
}

@article{zhong2026noise,
  title={From Noise to Intent: Anchoring Generative VLA Policies with Residual Bridges},
  author={Zhong, Yiming and He, Yaoyu and Yang, Zemin and Tian, Pengfei and Huang, Yifan and Huang, Qingqiu and Zhu, Xinge and Ma, Yuexin},
  journal={arXiv preprint arXiv:2604.21391},
  year={2026}
}

@article{li2026consisvla,
  title={Consisvla-4d: Advancing spatiotemporal consistency in efficient 3d-perception and 4d-reasoning for robotic manipulation},
  author={Li, Wei and Liu, Jizhihui and Yixing, Li and Tong, Junwen and Shao, Rui and Nie, Liqiang},
  journal={arXiv preprint arXiv:2605.05126},
  year={2026}
}

@inproceedings{feng2026spatial,
  title={Spatial-aware vla pretraining through visual-physical alignment from human videos},
  author={Feng, Yicheng and Zhang, Wanpeng and Wang, Ye and Luo, Hao and Yuan, Haoqi and Zheng, Sipeng and Lu, Zongqing},
  booktitle={Proceedings of the IEEE/CVF Conference on Computer Vision and Pattern Recognition},
  pages={712--723},
  year={2026}
}

@article{zhen20243d,
  title={3d-vla: A 3d vision-language-action generative world model},
  author={Zhen, Haoyu and Qiu, Xiaowen and Chen, Peihao and Yang, Jincheng and Yan, Xin and Du, Yilun and Hong, Yining and Gan, Chuang},
  journal={arXiv preprint arXiv:2403.09631},
  year={2024}
}
